\documentclass[conference,journal,transmag]{IEEEtran}

\usepackage[pdftex]{graphicx}
\usepackage{amsmath}
\usepackage{eqparbox}
\usepackage{tabularx}
\usepackage{mathtools}
\usepackage{tabu}

\begin{document}

\title{Early Diagnosis of Pneumonia with Deep Learning}

\author{\IEEEauthorblockN{Deniz Yagmur Urey\IEEEauthorrefmark{1},
Can Jozef Saul\IEEEauthorrefmark{1,2}, and
Can Doruk Taktakoglu\IEEEauthorrefmark{1}}
\IEEEauthorblockA{\IEEEauthorrefmark{1}Robert College of Istanbul, Istanbul, Turkey}
\IEEEauthorblockA{\IEEEauthorrefmark{2}Koc University Artificial Intelligence Laboratory, Istanbul, Turkey}

\IEEEauthorblockA{\IEEEauthorrefmark{1}ureden.20@robcol.k12.tr,\,\IEEEauthorrefmark{1,2}saucan.20@robcol.k12.tr,\,\IEEEauthorrefmark{1}takcan.19@robcol.k12.tr}}

\maketitle

\IEEEdisplaynontitleabstractindextext

\begin{abstract}
\emph{$\textbf{Abstract---}$}Pneumonia has been one of the fatal diseases and has the potential to result in severe consequences within a short period of time, due to the flow of fluid in lungs, which leads to drowning. If not acted upon by drugs at the right time, pneumonia may result in death of individuals. Therefore, the early diagnosis is a key factor along the progress of the disease. This paper focuses on the biological progress of pneumonia and its detection by x-ray imaging, overviews the studies conducted on enhancing the level of diagnosis, and presents the methodology and results of an automation of x-ray images based on various parameters in order to detect the disease at very early stages. In this study we propose our deep learning architecture for the classification task, which is trained with modified images, through multiple steps of preprocessing. Our classification method uses convolutional neural networks and residual network architecture for classifying the images. Our findings yield an accuracy of 78.73\%, surpassing the previously top scoring accuracy of 76.8\%.
\end{abstract}

\renewcommand\IEEEkeywordsname{Keywords}

\IEEEpeerreviewmaketitle

\section{Introduction}

Globally, 450 million get infected by pneumonia in a year and 4 million people die from the disease. 1 million people each year have to seek care from hospitals and 50 thousand people die from the disease \cite{cdn} in the United States of America. The numerical difference between the infection rates and death rates show how crucial the early diagnosis of the disease is. Pneumonia is an inflammatory response in the lung sacs called alveoli. It’s often caused by bacteria, viruses, fungi and other microbes. As the germs reach the lung, white blood cells act against the germ and inflammation occurs in the sacs. Thus, alveoli get filled with pneumonia fluid and this fluid causes symptoms like coughing, trouble in breathing and fever. If the infection isn’t acted upon during the early periods of the disease, pneumonia infection can spread throughout the body and result in the death of the individual, as a result of the inability to exchange gas in the lungs. 

Today, one of the most conventional medical techniques used to diagnose the disease is chest x-ray. As the concentrated beam of electrons, called x-ray photons, go through the body tissues, an image is produced on the metal surface (photographic film). During diagnosis, expert radiologists correspond white spots on the image to infiltrates identifying an infection, and white areas to the pneumonia fluid in the lungs. However, the limited color scheme of x-ray images consisting of shades of black and white, cause drawbacks when it comes to determining whether there’s an infected area in the lungs or not. This is due to the fact that the high intensity of white wavelength occurs on the photographic film when the fluid in the lungs is high enough to be considered as a dense and solid tissue. In other words, the transition from an air filled tissue (normal state of lungs), which is seen in darker shades, to a dense tissue, requires the sufficient amount of fluid to shift the color scheme to lighter colors. This means that for an x-ray film to be considered as pneumonia, the disease must be in its later stages. Thus, the early detection of pneumonia is restricted due to the limited color scheme of x-ray imaging. 

\begin{figure}[ht!]
\centering
\includegraphics[width=3.35in]{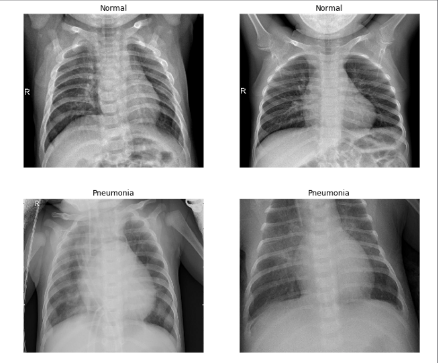}
\caption{X-ray Images with and without Pneumonia}
\label{xray_all}
\end{figure}

Another drawback for the early diagnosis of pneumonia is the human-dependent detection. Expert radiologists need to have sufficiently trained eyes in order to be able to differentiate between the heterogeneous color distribution of air while flowing in the lungs. This may be seen in different colors on the x-ray image taken, yet not be the dense pneumonia fluid. Thus, it's highly significant for a radiologist to be able to tell whether if the white spots on the x-ray film actually correspond to the fluid itself. As a result of the error margin of the human eye, there are many cases where the radiologists fail to make the correct diagnosis. In both cases, whether if it's a false positive or false negative diagnosis, it has substantial impacts on the human body. Therefore, computational methods in the diagnosis step of the disease are reliable in terms of consistency. In fig  \ref{xray_all}, different images with and without pneumonia can be seen (\cite{erthal}). The imperceptibility of the healthy versus the pneumonia images can also be witnessed, which portrays the need of well-trained eyes in order to be able to differentiate.

There has been previous studies done regarding pneumonia detection with chest x-rays via machine learning with the use of heat maps \cite{chexnet}, which are images or maps representing the varying temperature or infrared radiation recorded over an area or during a period of time, and differentiation of pulmonary pathology, which is the subspecialty of surgical pathology which deals with the diagnosis and characterization of neoplastic and non-neoplastic diseases of the lungs from normal by using computerized lung sound analysis \cite{clcsa}. Moreover, diagnosing p.carinii pneumonia, which is caused by fungi, with the examination of induced sputum and with indirect immunofluorescence \cite{carnii} has been used as a method for its specific detection. 

Aside from using the conventional x-ray imaging, diagnosing lower respiratory tract infection with techniques such as bronchoalveolar lavage, a medical procedure in which a bronchoscope is passed through the mouth or nose into the lungs and fluid is squirted into a small tube, lung biopsy \cite{biops}, which is a procedure performed to remove tissue or cells from the body for examination under a microscope, and using lung ultrasonography, which is a technique using echoes of ultrasound pulses to delineate objects or areas of different density in the body to detect neonatal pneumonia, has been done. Neonatal pneumonia is the lung infection in a newborn, which includes lung consolidation with irregular margins and air bronchograms, pleural line abnormalities, and interstitial syndrome \cite{inters}. There has also been previous studies done on the early detection of pneumonia. Among the various other methods used by different studies, this paper is the first one to present automations of various parameters on x-ray images, which can diagnose pneumonia at very early stages. 

While the mentioned conventional and radiological methods might be effective, our study presents a deep learning approach to this pneumonia classification. Looking at the state of art, there has been two previous similar experimentations on this task. The initial one (\cite{dnn1}) uses Long Short Term Memory (LSTM) architectures for finding interdependencies among the X-ray data. While their study focuses on 14 interdependent diseases, our study focuses merely on Pneumonia. However, due to their experimentation for extracting 14 different diseases with one model, they have merely been able to reach an accuracy of 71.3\%. Furthermore, LSTM uses multiple images for classifying a single image, whereas our proposed experimentation and model only need pre trained neural network weights for classifying images one by one. Additionally, our accuracy upon experimentation yielded 78.73\%, which uses the same dataset.

The second experimentation was conducted in Stanford University Machine Learning Laboratory (\cite{chexnet}). Their experimentation was conducted with similar means to ours. They used a 121 layer convolutional network for feature map acquisition, alongside with statistical methods (standard deviation and mean calculation) for image preprocessing. In our experiment we use three convolutional layers, yielding a more efficient and a computationally less costly training process. Our preprocessing methods are similar to real life applications, unlike statistical means that might be ineffective when wide range of data is present. Finally, our proposed architecture yields an accuracy of 78.73\%, while their study yielded an accuracy of 76.8\%.

Other than the above mentioned papers on pneumonia classification, Chest X-ray images have been widely subjected to experimentations with convolutional neural network architectures, as well as other image classification techniques. Bone structures were segmented within a paper (\cite{conv3}). This paper presents a segmentation method which utilizes additional steps after the classification algorithm. As a regular Convolutioal neural network classifies the image as a whole, such segmentation methods utilzie pixelwise classification, which, in the end, applies a deconvolutional layer for classifying each pixel one by one and eventually seperating different objects within an image, bones being the most prevalent ones for the mentioned task. 

In another research (\cite{conv4}) aiming to conduct early detection not for pneumonia but for thorax disease through weekly classifications with convolutional neural networks. This paper successfully detects patterns for patients who have thorax disease or one that might have the mentioned disease. Yet, has no activity on pneumonia classification.

In this study we present a novel method for classifying pneumonia existence in an x-ray image. We propose a two-step image processing before training our deep learning model, in order for making the features of an x-ray image clearer and explicit for easing the classification process. We, then, execute a convolutional neural network followed by a residual neural network for the classification process. This paper will first explain the methodology in our experimentation, followed by the discussion of the results at hand.

\section{Methods}
\subsection{Data Preprocessing}
\begin{figure}[ht!]
\centering
 \includegraphics[width=2.95in]{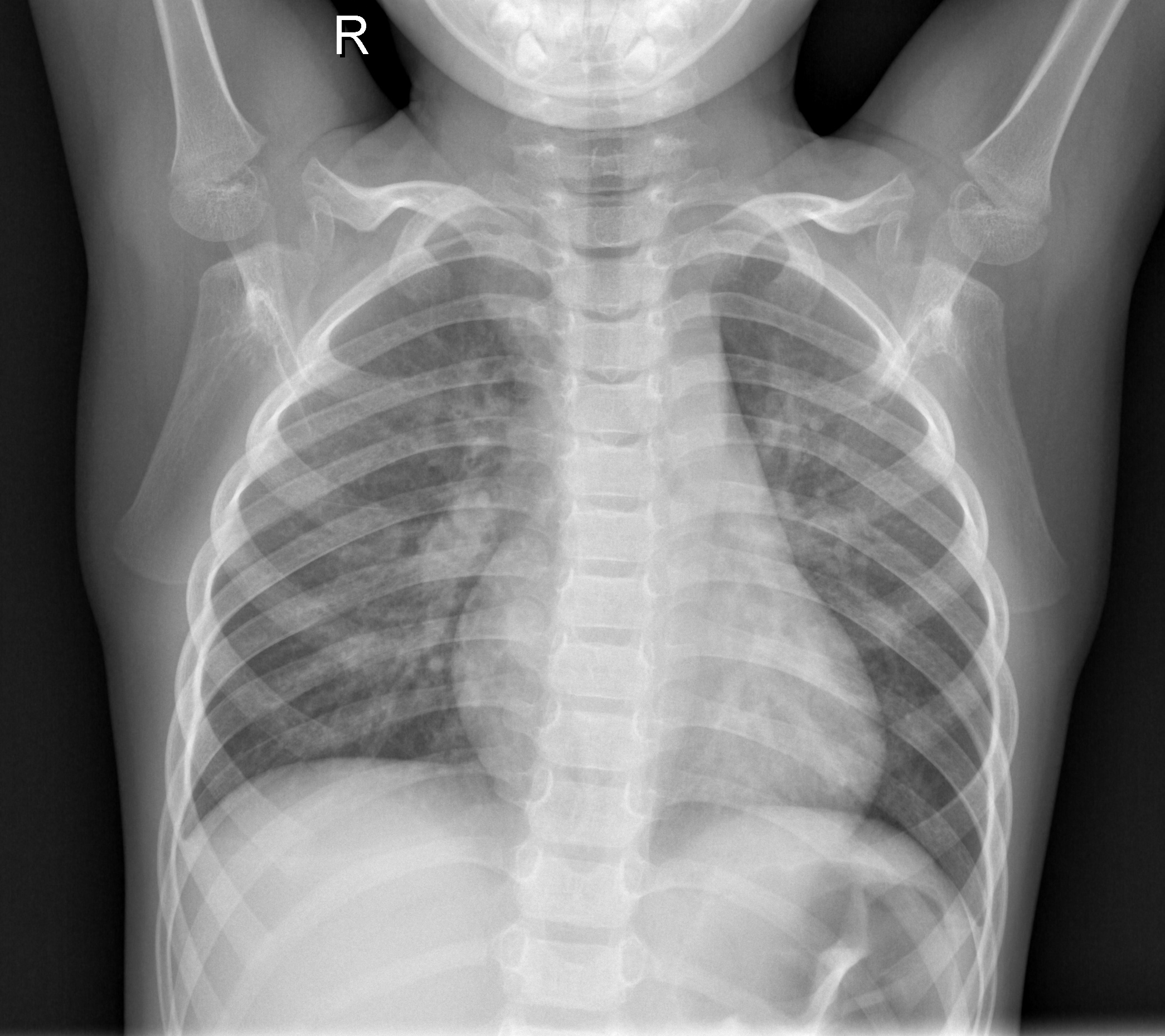}
\caption{Original X-ray Image}
\label{xray}
\end{figure}

The dataset was released on a public website, kaggle.com. The dataset was released by the Radiological Society of North America, which specified an x-ray image’s identity and whether if pneumonia is present in the x-ray data. We have used approximately 3 thousand images for image training and approximately 1 thousand images for image testing. All the images are from x-ray, which has limited color space, therefore, on the RGB scale, the image doesn’t show difference on the edges or in the parts where certain features might be detected. Therefore, we have applied certain color modifications in our image preprocessing. Our experimental methodology involves three different image processing techniques: increment in contrast, widening of the image color space and artificially lighting the image (increase in brightness). The fig \ref{xray} is the original figure for reference.

\begin{figure}[ht!]
\centering
\includegraphics[width=2.95in]{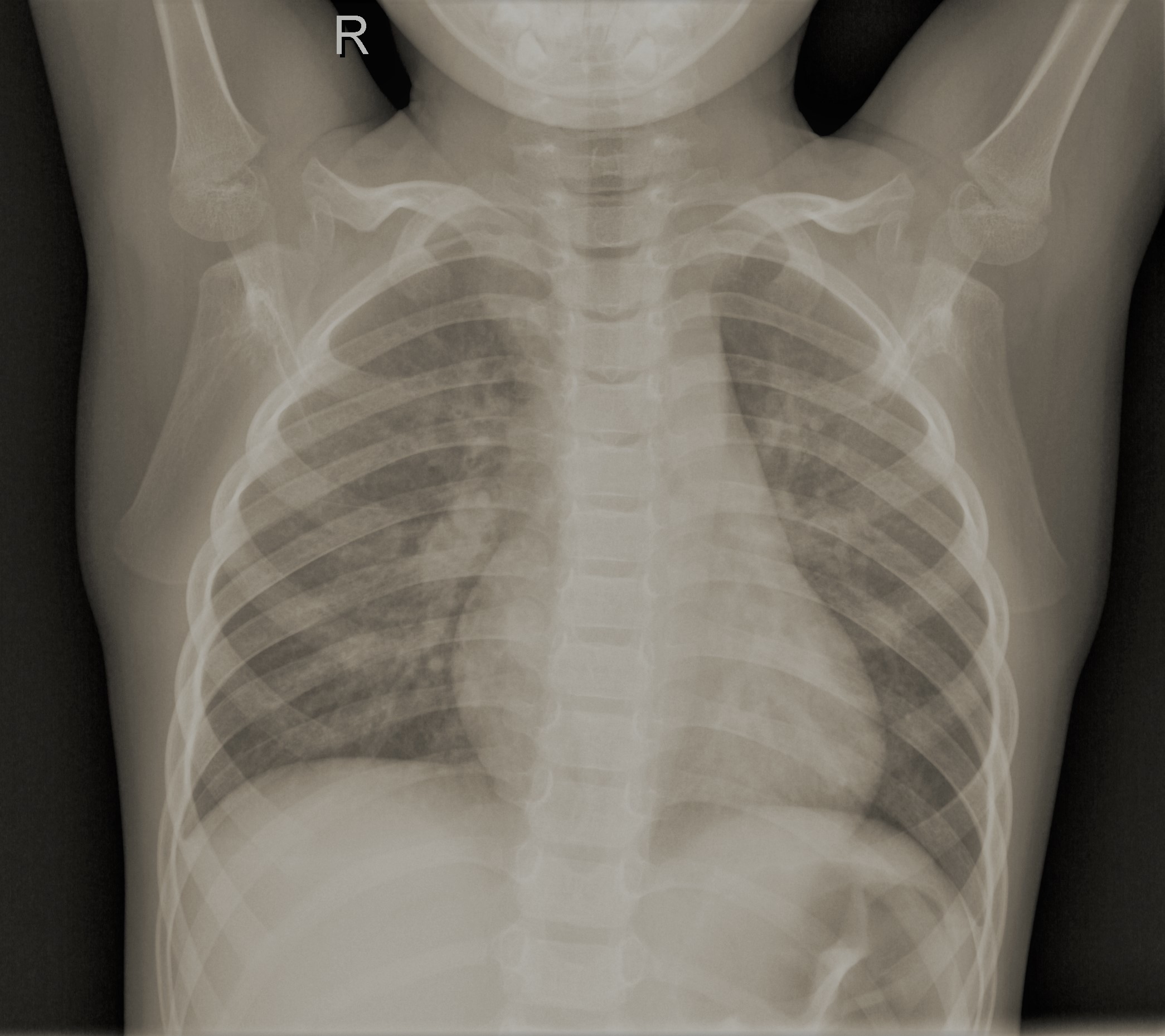}
\caption{X-ray Image with Modified Lightening}
\label{xray_l}
\end{figure}

The initial experimentation technique was image lighting and increment in brightening. This technique was utilized with the notion that even the professional doctors examine an x-ray image under light. Therefore, applying the same effects artificially might be an essential feature of the extraction step for the classification task. Increment in brightness is executed through parsing every single pixel of an image and then increasing their respective Red Green Blue values by a constant. A sample of this type of preprocessing can be visualized in fig \ref{xray_l}.

\begin{figure}[ht!]
\centering
\includegraphics[width=2.95in]{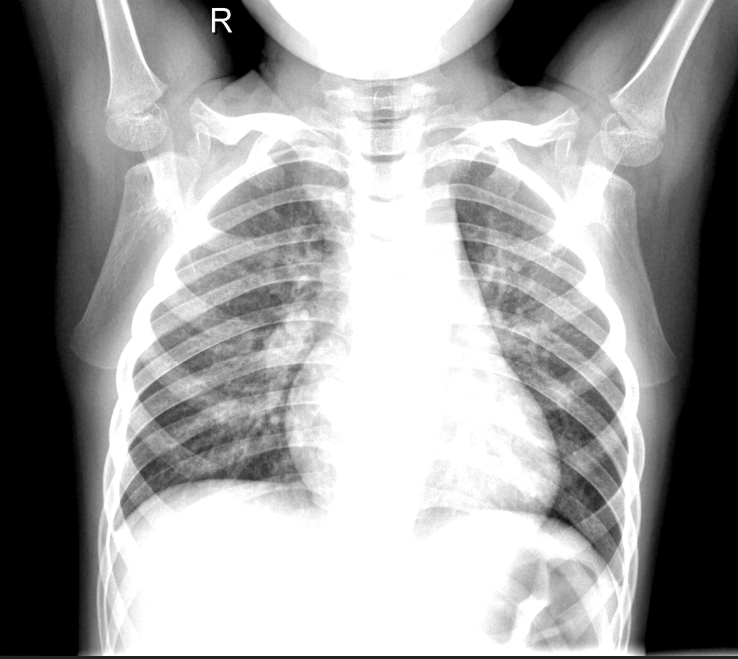}
\caption{X-ray Image with Modified Contrast}
\label{cont}
\end{figure}

The second technique we used was increment in image contrast, which is similar to changing image brightness. Increasing the image contrast makes the edges more solid and certain regions more visible. This technique was used as the x- with its original color scheme and doesn’t reflect the features, therefore, becoming essential for emphasizing certain parts of the image. Increment in image contrast can be described with the eq. \ref{contrast}. A sample of this modification can be found in fig \ref{cont}.

\begin{equation}\label{contrast}
g(i, j) = \alpha * f(i, j) + \beta
\end{equation}

Where $\alpha$ is the contrast, $\beta$ is image brightness and $i, j$ are the coordinates of respective pixels in an image.

\begin{figure}[ht!]
\centering
\includegraphics[width=2.95in]{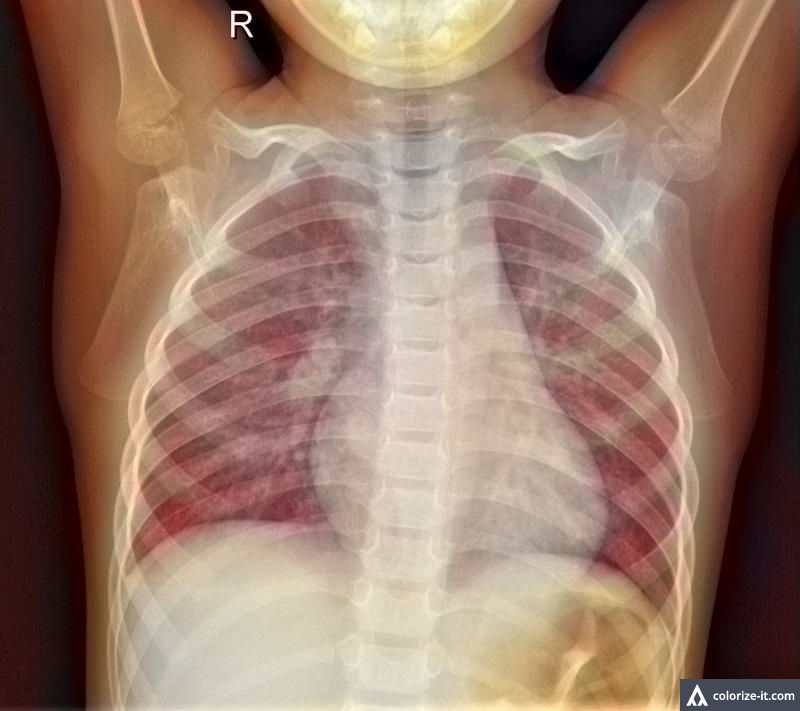}
\caption{X-ray Image with Expanded Color Scheme}
\label{color}
\end{figure}

The third technique was the expansion of the color scheme. On the execution side of this technique, the average R, G and B values are found among the images. Then, all the respective RGB values are multiplied with the mentioned average value for expanding, increasing the overall values and yielding a colorized version of the image. This technique was used in order to make the features clearer while classifying the image. A sample of the colorized version can be found in fig \ref{color}.

\begin{figure}[ht!]
\centering
\includegraphics[width=2.95in]{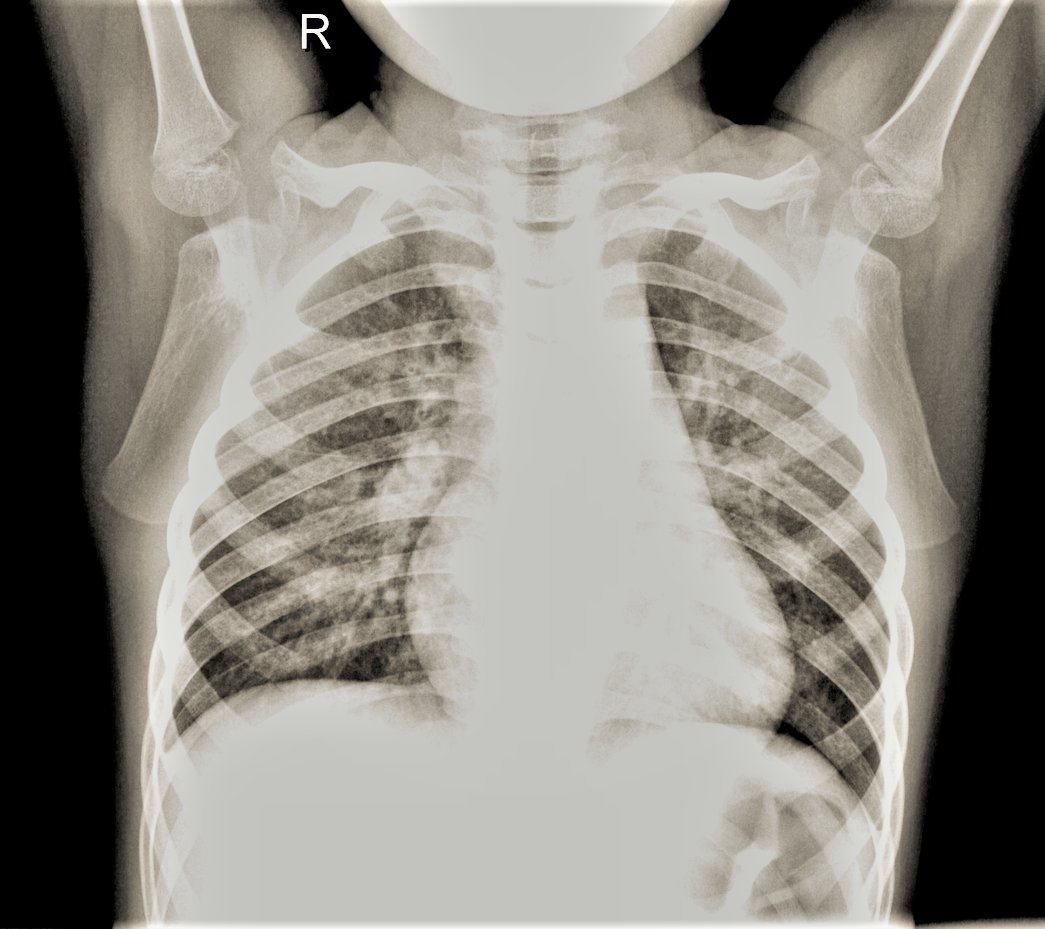}
\caption{X-ray Image with combined pre-processing methods applied}
\label{final_vrs}
\end{figure}

The finalized version of the image after our pre-processing pipeline can be found in fig \ref{final_vrs}. The image we created enables certain details to emerge so that the convolutional neural network can better detect any differences that indicate either an image is pneumonia or not.

\subsection{Classification}
This section will elaborate on the classification algorithms that were used throughout the experimentation process.
\\
\subsection{Convolutional Neural Network}
Convolutional Neural Networks are powerful tools for recognizing local patterns in data samples. As interrelated weight data is present in data samples, CNNs are suitable architectures for the classification task.

The following paragraphs will explain how our experimented CNN architecture functions. As mentioned, a CNN detects local patterns in an input by creating feature maps. Feature maps are created through conducting element wise multiplication with our kernel and the slided area of the input value. Then all the values are summed, yielding a result for the feature map. The mentioned process’s two dimensional version is summarized in fig \ref{feat}. Feature maps formulation is an essential step for classification as it manages to extract the significant portion of the information within an image while eliminating the unnecessary ones. The two dimensional process in fig \ref{feat} is conducted for every single layer (R, G, B) of an image and after its completion the image's layers are concatenated for the next step, classification through an artificial neural network.

\begin{figure}[ht!]
\centering
\includegraphics[width=2.85in]{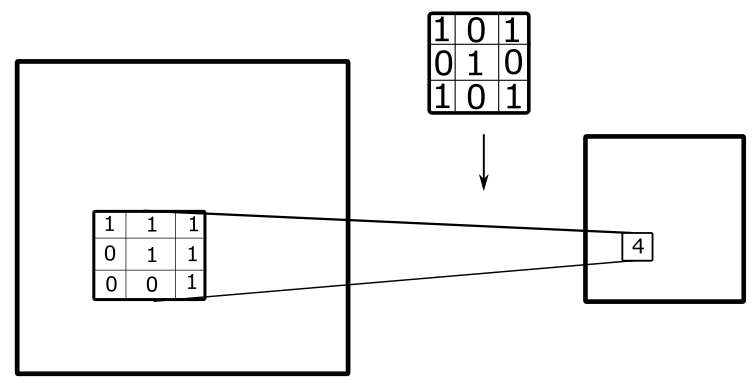}
\caption{Feature Map Formulation in A Convolutioanl Neural Network}
\label{feat}
\end{figure}

After a feature map is captured, a nonlinear function is applied, converting every negative value to 0 and maintaining all positive values as are. The mentioned function can be described with the Rectified Linear Unit (ReLU) function, eq. \ref{relu}. Non-linearity is utilized here since the data at hand can’t be merely described with linear functions, and therefore, non-linearity is crucial for detecting patterns in our data.

\begin{equation} \label{relu}
y = max(0, x)
\end{equation}

Then, pooling application is applied, which comes in variations of maximum, average and sum pooling. In our architecture, max pooling is utilized as it has been found more effective in previous studies \cite{pool}. Max pooling reduces the dimensions of the feature map while maintaining the most important identity values through sliding kernels over the rectified feature map and merely capturing the highest values.  Pooling is applied for making the data more manageable with less parameters, as the dimensions are reduced. For the following steps, the current output is flattened, converted to one long vector, which will be crucial for the classification algorithms. Flatting is applied for converting the data to a more manageable version within the classification algorithm, artificial neural network, as it intakes the flattaned data as input. The flattening process can be visualized in \ref{flat}.
 
\begin{figure}[ht!]
\centering
\includegraphics[width=2.95in]{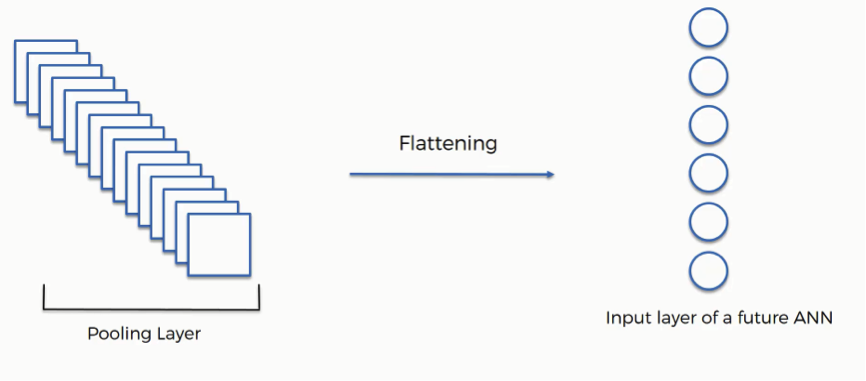}
\caption{Flattening after the Convolutional Steps}
\label{feat}
\end{figure} 
 
After this point, the network obtained the feature map of the input value, which will proceed with a regular feed forward back propagation neural network. The following paragraphs of this section will elaborate on the feed forward artificial neural network architecture.
 
Artificial neural networks (ANNs) are comprised of layers, which are comprised of multiple perceptrons. The perceptrons are fed with input values coming from the previous layers, which can be another hidden neuron layer or an input layer. The perceptron equation can be found in eq. \ref{perceptroneq}. 

\begin{equation} \label{perceptroneq}
y = \phi({\sum_{i=1}^{n} W_i* x_i})
\end{equation}

\begin{figure}[ht!]
\centering
\includegraphics[width=2.15in]{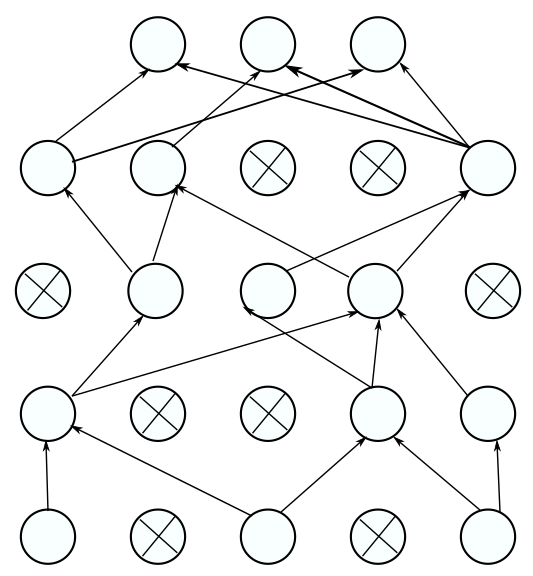}
\caption{Dropout Operation Visualization}
\label{do}
\end{figure}

Where $\phi$ represents an activation function, $x$ represents the input value and $w_i$ represents the layer weights. The activation functions utilized in our experimentation are either hyperbolic tangent, eq. \ref{tanh}, or ReLU (Rectified Linear Unit) function (eq. \ref{relu}) and the sigmoid function (eq. \ref{sigmoid}), for binary classification. Really high number of neurons or layers will enforce the network to memorize the dataset, leading to an inability to make accurate predictions in testing. Therefore, we use dropouts for eliminating the issue of overfitting. Dropout is a regularization technique preventing the network from memorizing a specific dataset and rather enabling it to adapt to variant inputs \cite{dropout}. Dropout visualization can be found in fig \ref{do}. Additionally, batch normalization is applied after certain layers for normalizing the output values before the activation functions are applied. After all the hidden layers are passed, softmax function (eq. \ref{softmax})  is applied for attaining a probability map for the output values. Then  - during the network training period - a loss value is calculated, which we used various functions during the experimentation, then the network is backpropagation with an optimizer function for updating layer weights, which were randomly initialized. 

\begin{equation}\label{tanh}
y = \frac{1-e^{-2x}}{1+e^{-2x}}
\end{equation} 

\begin{equation}\label{sigmoid}
 h_ \theta (x) =  \frac{\mathrm{1} }{\mathrm{1} + e^- \theta^Tx }
\end{equation}

\begin{equation}\label{softmax}
p_c = \frac{e^{W_{r+b}}}{\sum_{i=1}^{L} e^{W_{i^{r+b_i}}}}
\end{equation}
 
For our network, we used three convolutional layers and max pooling. Alongside hyperparameter tuning, we added a dropout layer after the Softmax function for the prevention of overfitting. Dropout function randomly drops out a specified amount of neurons from the neural network. We used the Adam optimizer for our network, which enables rapid convergence compared to other optimizers \cite{adam}. Binary cross entropy function was used as our loss function. Sequentially, filter sizes of 3 and 4 were used with a dropout rate of 0.4. The training data was trained with 120 epochs, a batch size of 40 and a learning rate of 0.001.

\subsection{Residual Neural Network}
\begin{figure}[ht!]
\centering
\includegraphics[width=2.15in]{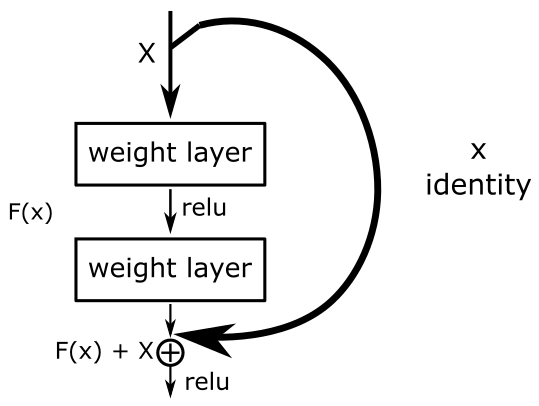}
\caption{Residual Neural Network Visualization}
\label{resnet}
\end{figure}

Our alternative model for improving the prediction side of the network was the residual neural network architecture \cite{resnetc} . This uses residual block for maintaining identities (as seen in fig. \ref{resnet}), created by activation functions (hyperbolic tangent function for our architecture), throughout the network.  The mentioned ability is implemented through the summation of the result of a linear function with the result of the prior activation function. Only one residual block was present in our network. We used a 9 layered ResNet architecture in our experimentation after fine tuning the network. As also presented in the original Residual Neural Network Paper \cite{resnetc}, such architecture's tuning generally indicates higher accuracies when it has more layers. The original paper indicates an increasing accuracy on COCO dataset when layers are increased from eleven to hundred and twenty one. However, with the present computational power we have, a nine layered network was the top result we were able to formulate within a short time period.

\section{Results and Discussion}
The results from our experimentation can be found in table \ref{tab:res}.

\begin{table}[h]
\caption{Experimental Results}
\begin{tabular}{|c|c|l}
\cline{1-2}
\textbf{Network}                                           & \textbf{Accuracy} &  \\ \cline{1-2}
CheXnet (previously proposed model)                        & 76.80\%           &  \\ \cline{1-2}
CNN with Unmodified Input                                  & 63.74\%           &  \\ \cline{1-2}
CNN with Expanded Color Scheme                             & 65.42\%           &  \\ \cline{1-2}
CNN with Increased Contrast                                & 69.92\%           &  \\ \cline{1-2}
CNN with Lightened Image on Increased Contrast             & 75.65\%           &  \\ \cline{1-2}
CNN with Lightened Image on Increased Contrast with ResNet & 78.73\%           &  \\ \cline{1-2}
\end{tabular}
\label{tab:res}
\end{table}

The defintion in eq. \ref{accuracy} will be used throughout the discussion of the results.

\begin{equation}\label{accuracy}
Accuracy = \frac{\text{number of correct results by the network}}{\text{total tests done by the network}} \times 100
\end{equation}

Our results show improvement in performance over different modifications. Initially, the base model was able to reach approximately the same value in accuracy as other papers were able to do. Then with an expanded color scheme, more features were made clear in the image, yielding a higher accuracy for the classification task. However, there wasn’t a major difference relative to the base model as all the RGB values were multiplied by a calculated constant. As there isn’t a major increment due to constant not being variables, features weren’t very cleared out in this classification task.

CNN with increased contrast also shows an improvement relative to the base with a higher value in performance. Increased contrast was crucial for having the edge-like features more clear in the image. There is also an increment relative to the version with expanded color scheme which is, as mentioned, due to the method of making modifications on the images: while expansion of color scheme uses a constant, increment in contrast is variant throughout an image, depending on every single pixel.

The highest accuracy with a standard artificial neural network on classification side was acquired from images modified with artificial lighting on top of increased contrast. As expected, this model yielded the highest accuracy considering its similarity to real life classification process. Increment in contrast, as the previous experiment, made certain features mode clear while change in lightning further emphasized on certain features through change in brightness on certain parts of the image, depending on the pixel’s respective RGB values. Then, the addition of residual architecture with increased number of hidden layers has also improved the performance rate, which is due to the maintenance of scalar identity and normalization of batch values through addition, rather than feature scaling.

In comparison to the previous state of art, our results surpass the previous studies in terms of the accuracy. While CheXNet had an accuracy of 76\% in classification, out network has been able to classify with an accuracy of 78.73\% in the fine tuned version. Our primary contributions in this comparison is increment in classification accuracy, decrease in computational time as CheXNet \cite{chexnet} was able to reach such a performance with a 121-layer convolutional neural network, while our experiment used 3 convolutional layers, and a new state of feature extraction for the classification task. Usage of 121 convolutional layers causes a lengthy time for training the network, while parameters of computational power are kept the same. However, the testing duration is merely varied due to change in computational power. While CheXnet used statistical means such as standard deviation for feature extraction before the classification, our model used modification on contrast and lightning for obtaining a higher accuracy.
Our model reaches an F score of 45.79\%, whereas the biological, radiological methods reach an F-Score of 38.7\% in their classification algorithms \cite{chexnet}. Therefore, our classification methods show a greater accuracy than the conventional methods.

\section{Conclusion}
In this study, we present a novel method for classifying an X-ray image on its possibility of exhibiting pneumonia in the early stages of the disease. We experiment with three different preprocessing techniques: increase in colorspace, increase in contrast and artificially lightening of the image. We've used multiple combinations of preprocessing techniques with various networks. In our final experimentation, we combined increment in contrast and lightening methods for incorporating both of the feature extraction techniques. We used a convolutional neural network approach for obtaining feature maps of the preprocessed X-ray images. Then, we experimented with two different classification methods. The first method was an artificial neural network while the second was the ResNet architecture, which yielded a higher accuracy. Our most accurate experimentation model classifies the images with a 78.73\% accuracy, surpassing the previously top scoring value from CheXnet \cite{chexnet}. Overall, we target the current drawback in medical diagnosis of pneumonia by the human high, and propose an alternative and more accurate way of diagnosing the disease with automation. Moreover, we target the limits caused by the gray scale of x-ray imaging, preventing the early diagnosis of the disease. Our study presents an efficient algorithm with a high performance for this classification task and can be improved through object detection algorithms for extracting the region with pneumonia. YOLO and SSD algorithms might be effective for the localization of the pneumonia region, while different preprocessing methods may be needed for training the respective algorithms.

\end{document}